\documentclass{article}

\usepackage{arxiv}

\usepackage[utf8]{inputenc} 
\usepackage[T1]{fontenc}    
\usepackage{hyperref}       
\usepackage{url}            
\usepackage{booktabs}       
\usepackage{amsfonts}       
\usepackage{nicefrac}       
\usepackage{microtype}      
\usepackage{lipsum}		
\usepackage{graphicx}
\usepackage{doi}
\usepackage[version=4]{mhchem}
\usepackage{amsmath,amssymb}

\title{Hyperdimensional computing: a fast, robust and interpretable paradigm for biological data}

\author{ \href{https://orcid.org/0000-0003-0903-6061}{\includegraphics[scale=0.06]{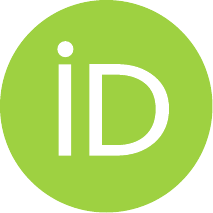}\hspace{1mm}Michiel Stock} \\
    Department of Data Analysis\\ and Mathematical Modelling\\
    Ghent University\\
    \texttt{michiel.stock@ugent.be} \\
	\And
	\href{https://orcid.org/0000-0000-0000-0000}{\includegraphics[scale=0.06]{orcid.pdf}\hspace{1mm}Dimitri Boeckaerts} \\
	Department of Data Analysis\\ and Mathematical Modelling\\
 Department of Biotechnology\\
    Ghent University\\
 \And
	\href{https://orcid.org/0000-0003-4564-1672}{\includegraphics[scale=0.06]{orcid.pdf}\hspace{1mm}Pieter Dewulf} \\
	Department of Data Analysis\\ and Mathematical Modelling\\
    Ghent University\\
     \And
	\href{https://orcid.org/0000-0002-2685-4130}{\includegraphics[scale=0.06]{orcid.pdf}\hspace{1mm}Steff Taelman} \\
	Department of Data Analysis\\ and Mathematical Modelling\\
    Ghent University\\
     BioLizard nv\\
 \And
	\href{https://orcid.org/0000-0001-7938-1675}{\includegraphics[scale=0.06]{orcid.pdf}\hspace{1mm}Maxime Van Haeverbeke} \\
	Department of Data Analysis\\ and Mathematical Modelling\\
    Ghent University\\
 \And
\href{https://orcid.org/0000-0003-2971-5539}{\includegraphics[scale=0.06]{orcid.pdf}\hspace{1mm}Wim Van Criekinge} \\
Department of Data Analysis\\ and Mathematical Modelling\\
Ghent University\\
 \And
\href{https://orcid.org/0000-0002-3876-620X}{\includegraphics[scale=0.06]{orcid.pdf}\hspace{1mm}Bernard De Baets} \\
Department of Data Analysis\\ and Mathematical Modelling\\
Ghent University\\
}

\renewcommand{\shorttitle}{Hyperdimensional Computing for Biological Data Analysis }

\hypersetup{
pdftitle={A template for the arxiv style},
pdfsubject={q-bio.NC, q-bio.QM},
pdfauthor={David S.~Hippocampus, Elias D.~Striatum},
pdfkeywords={First keyword, Second keyword, More},
}

\begin{document}
\maketitle

\begin{abstract}
Advances in bioinformatics are primarily due to new algorithms for processing diverse biological data sources. While sophisticated alignment algorithms have been pivotal in analyzing biological sequences, deep learning has substantially transformed bioinformatics, addressing sequence, structure, and functional analyses. However, these methods are incredibly data-hungry, compute-intensive and hard to interpret. Hyperdimensional computing (HDC) has recently emerged as an intriguing alternative. The key idea is that random vectors of high dimensionality can represent concepts such as sequence identity or phylogeny. These vectors can then be combined using simple operators for learning, reasoning or querying by exploiting the peculiar properties of high-dimensional spaces. 
Our work reviews and explores the potential of HDC for bioinformatics, emphasizing its efficiency, interpretability, and adeptness in handling multimodal and structured data. HDC holds a lot of potential for various omics data searching, biosignal analysis and health applications. 
\end{abstract}


\section{Introduction}

\section{Introduction}
Computational biologists and bioinformaticians collect, organize, process, and analyze large amounts of biological data to extract biological knowledge~\cite{gauthierHistory2019}. Parallel advances in biological data generation and computer science have further expanded the capabilities and usefulness of bioinformatics, proving immensely valuable for uncovering biological knowledge from sequence data. Today, bioinformatics is being transformed once again by deep learning (DL)~\cite{LeCun2015deeplearning} and its ability to handle complex, high-dimensional and multimodal data such as sequences and images, redirecting interest from earlier powerhouses such as kernel-based learning~\cite{scholkopfKernelMethodsComputational2004,greenerML2022}.
Prominently, the development and use of complex DL models such as AlphaFold~\cite{jumperAlphaFold2021}, ESMFold~\cite{linESM2023} and RoseTTAFold~\cite{baekRosetta2021} represent a paradigm shift in protein structure prediction. Moreover, DL is also leading to significant breakthroughs in other fields, such as protein design~\cite{yehDeNovo2023}, medical imaging~\cite{anaya2021overview} and drug discovery~\cite{liuDL2023}, with modest advances in fields such as systems biology and phylogenetic inference~\cite{sapovalCurrentProgressOpen2022}.
Curiously, there is a disparity between the fields in which the impact of DL can, to some extent, be explained by only the need for learning a mapping based on large datasets (e.g., protein structure prediction) versus the fields in which the problem settings involve complex combinations of structured data and information (e.g., multi-omics and phylogeny).

Two limitations currently hamper the utilization of DL models in bioinformatics~\cite{sapovalCurrentProgressOpen2022}.
Firstly, large connectionist models are often black boxes, while the explainability of models is an essential property for biologists, arguably more so than predictive performance.
For example, when medical practitioners use a model to aid in making a diagnosis or finding a treatment, they must understand why this conclusion was reached~\cite{hanChallengesExplainableAI2022}. 
Despite substantial advances in explainable machine learning~\cite{montavonMethodsInterpretingUnderstanding2018,liuModular2023}, DL methods still lack the clarity inherent in methods such as decision trees or logistic regression. 
Secondly, DL models are typically costly to train regarding the required data and the associated computational cost. Most DL methods are very data hungry -- with some notable exceptions, e.g., a recent RNA folding model trained on only 18 structures~\cite{Townshend2021geomRNA}.
Training a single competitive DL model may cost tens to hundreds of thousands of US dollars and has a high environmental cost regarding \ce{CO2} emissions~\cite{strubellEnergyPolicyConsiderations2019}.
Meanwhile, transfer learning and fine-tuning have emerged as approaches to circumvent large additional training costs~\cite{iman2023review}. Furthermore, developing efficient architectures and training protocols is an active area of research~\cite{hanEfficient2015, liuEfficient2018}.

This work explores the potential of hyperdimensional computing (HDC), sometimes called vector symbolic architectures (VSA), as an alternative learning and information processing paradigm for bioinformatics~\cite{Kanerva2009hyperdimcomp}. While abstract models of the brain inspire HDC and DL, they are very different. Rather than mimicking the hierarchical connectionist neural architecture, HDC is a conceptual model of how representations are stored in the human brain. Here, concepts are represented by high-dimensional vectors (i.e., $10,000$ dimensions or more), the eponymous hypervectors (HV). HDC uses a set of mathematical operations to combine and change the information stored in different vectors to create an associative memory, a database of concepts. Using a small set of mathematical operations, one can construct, process, combine, split or query the concepts in this database. For high-dimensional vectors, one can show that similarity metrics such as the cosine similarity or similarity based on the Hamming distance are extremely sensitive to detect related vectors. Rather than being based on exact, algorithmic computing, HDC uses a cybernetic form of computation~\cite{jaeger2023toward} where concepts are stored in a distributed fashion. Inferences are made by computing the similarity between query vectors and those stored in a memory, similar to how nearest-neighbor and other prototype methods work. Having many attractive characteristics that will be explored in the sections below, we believe HDC is a promising complementary paradigm to DL in bioinformatics with a wide range of applicability.

In the next section, we first outline the characteristic aspects of HDC. Then, we provide an accessible, though relatively comprehensive, introduction to HDC, including the different strategies of creating HVs, the basic operations and their intuition, how to represent the most commonly-used data types (numbers, vectors, sequences, graphs, etc.), and how learning is performed. Finally, we discuss the strengths and promising applications of HDC for bioinformatics and computational biology. Though HDC is considered an obscure topic in some circles, there exists a vast amount of exciting work, which we cannot hope to cover comprehensively. This paper should also serve as a general introduction for computational life scientists. Throughout, we point to other work that is more specific or broader in scope. 

\section{The nature of hyperdimensional computing}\label{sec:nature}
\begin{figure}[ht!]
    \centering
    \includegraphics[width=\textwidth]{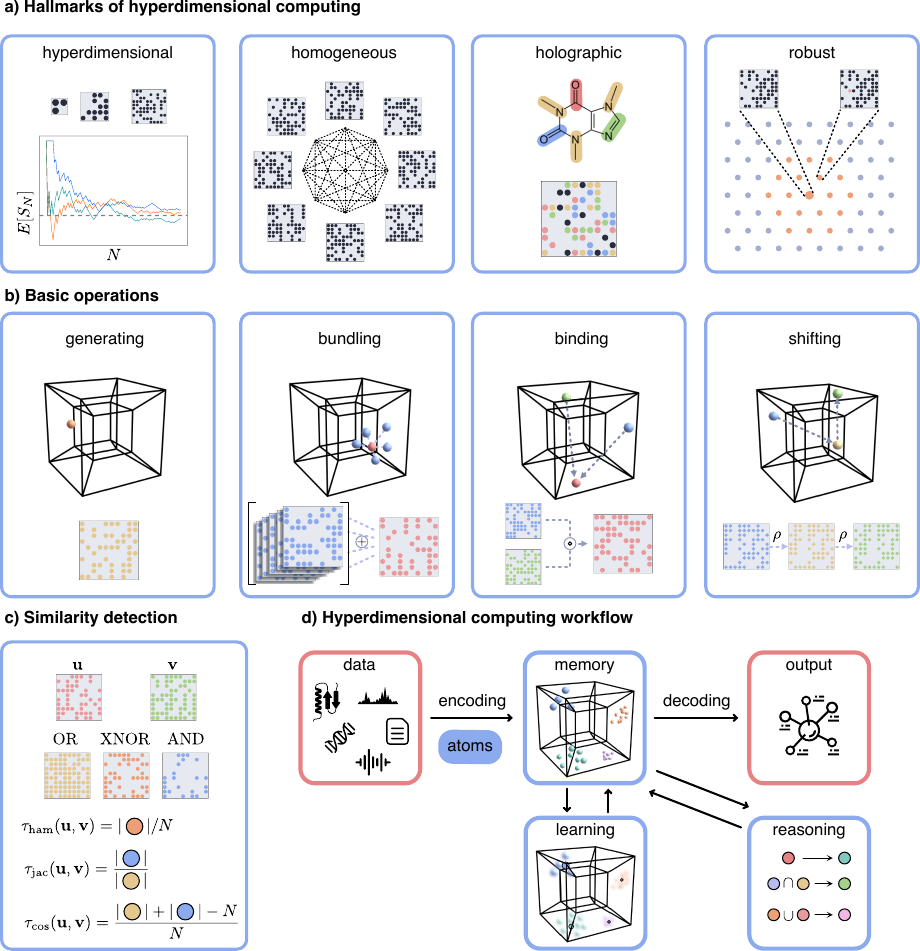}
    \caption{a) The hallmarks of hyperdimensional computing (HDC). Hypervectors (HVs) work reliably due to their large dimensionality $N$ (i.e., the Law of Large Numbers states that element-wise properties $S_N$, such as the fraction of positive elements, converge to their expected value for large $N$), and the space is very homogeneous (e.g., most HVs are approximately equidistant). The information about an object is encoded holographically, and the information is robust to random errors. b) Overview of the elementary operations of hyperdimensional computing (HDC): generating, bundling, binding, and shifting. c) Similarity is computed based on element-wise comparisons. d) General HDC workflow, based on Thomas and Rosing~\cite{Thomas2021theoryofHDC}, where red boxes indicate the data space and blue boxes indicate operations in the hyperdimensional space.}
    \label{fig:operations}
\end{figure}

The hyperdimensional computing emerged in the nineties and has recently seen a surge of interest in the machine learning community~\cite{kanervaHyperdimensionalComputingAlgebra2022,kleykoSurveyHyperdimensionalComputing2022}.
It originates from a broader range of computational models that, by implementing an efficient binding operation to combine different sources of information, attempt to combine the benefits of so-called old-fashioned symbolic AI with the more modern connectionist and data-driven machine learning approaches. After the introduction of tensor product representations~\cite{smolensky1990tensor}, many similar models, such as holographic reduced representations~\cite{plate1995holographic}, binary spatter codes~\cite{kanerva1996binary} and multiply-add-permute~\cite{gayler1998multiplicative} have been suggested.  
These models rely on the same mathematical properties to represent entities in high-dimensional spaces, leading to the term `hyperdimensional'.
Today, various models exist to construct so-called hypervectors, for example, using binary, real-valued, or complex elements. For an elaborate overview, we refer to~\cite{kleykoSurveyHyperdimensionalComputing2022, schlegelComparisonVectorSymbolic2022}.
Rather than the specific choice of the values in the hyperdimensional representations, we identify four hallmarks that distinguish hyperdimensional computing from other approaches (Figure~\ref{fig:operations}a). These are:
\begin{enumerate}
    \item \textbf{hyperdimensional}: the HVs live in
    a very high-dimensional space, large enough such that random elements can be seen as distinct and dissimilar from one another;
    \item \textbf{homogeneous}: the vast majority of HVs all have highly similar properties: they have (approximately) the same norm, are all equally (dis)similar to one another and have the same dimensionality, even if they embed more complex information, etc.;
    \item \textbf{holographic}: the information encoded in a HV is distributed over its many dimensions; no specific region is more informative than another for a specific piece of information;
    \item \textbf{robust}: randomly changing a modest number of the elements does not substantially change a HV's meaning.
\end{enumerate}

These hallmarks characterize the nature of the hypervectors, and together with a well-chosen set of mathematical operations, they allow one to encode complex structures such as amino acids, genes, gene regulatory networks, proteins, or whole genomes. As a general guide, information is preserved using \textit{similarity}: similar entities or complex constructions have similar HV representations. A sensitive similarity measurement, such as one based on the Hamming distance or the Jaccard similarity, is vital for inference and querying. 

The first vital aspect of the power of HDC is the \textbf{high dimensionality}, typically $10,000$ as a guideline. It leads to an astronomical representational power for complex objects such as genes or networks: two randomly selected vectors will likely be dissimilar, allowing them to store information independently. Hyperdimensionality also leads to robust systems, a phenomenon known in mathematics, statistics and physics as \emph{the blessing of high-dimensionality}. For example, in statistical physics, systems with many degrees of freedom lead to robust behavior that can be described by emergent properties such as pressure~\cite{gorbanBlessingDimensionalityMathematical2018}. 
Different data types are encoded in the same form of hypervectors, a property we call \textbf{homogeneity}.
One cannot tell whether a specific HV encodes a more complex concept, such as a protein, or an atomic concept, such as an amino acid.
In HDC, individual elements of the HVs cannot be linked to specific information about an entity. Instead, all elements contribute slightly to representing all the properties at once. This distributed representational property is called \textbf{holographic}. Storing a little bit of all the properties in every element is the basis for homogeneously constructing complex objects. This property is in marked contrast with, for example, TF-IDF word embedding vectors (each element corresponds to the occurrence of a word in a particular text) or molecular fingerprints (elements correspond to the occurrence of specific subgroups of the molecule).
Finally, HDC is \textbf{robust to noise} because of the above properties. Due to its holographic nature, each element comprises the same information but in a differently corrupted way, such that hyperdimensionality 
ensures a representation that is inherently robust to corruption. This will allow for the construction and manipulation of complex entities without too much loss of information. In essence, computing similarities of large, randomly initialized vectors can be seen as approximating expected values, which are preserved under unbiased corruption, i.e., noise.

\section{A gentle introduction to HDC}\label{sec:introhdc}

\subsection{Computing with HVs}

The basic operations needed for HDC are remarkably simple.
In brief, they hinge on four operations to manipulate and extract the information in the vectors (Figure~\ref{fig:operations}b,c). These are
\begin{itemize}
    \item generating new HVs from scratch;
    \item combining a set of HVs into a new HV that is similar to all;
    \item using one or more HVs to generate a new one that is dissimilar to its parent(s);
    \item comparing two HVs to detect whether they are more (dis)similar than expected by chance.
\end{itemize}
This work will limit the discussion to the most basic, generally-used cases. For an exhaustive overview, we refer to survey papers such as~\cite{kleykoSurveyHyperdimensionalComputing2022}. We also refer to~\cite{schlegelComparisonVectorSymbolic2022}, who compare eleven different HDC architectures in-depth and \cite{Thomas2021theoryofHDC} for theoretical analysis. As a running example, we will use the encoding of amino acid sequences to explain the operations.

\subsubsection{Hypervector generation}

Firstly, one has to fix the nature of the HVs, e.g., whether to work with binary ($\{0,1\}$), bipolar ($\{-1,1\}$), sparse, or real-valued vectors. 
One needs to define a function for these types to generate new \emph{atomic vectors}.
Atomic vectors represent the basic building blocks of the entity of interest.
For example, protein sequences consist of amino acids, DNA sequences of nucleotides and protein networks of proteins.
These are atomic in the sense that they are, within their context, not composed of simpler substructures.
In practice, generation can be done by filling a vector with i.i.d.\ (pseudo-)random numbers of the appropriate type, e.g., Booleans drawn from a Bernoulli distribution, $-1$ and $1$ drawn from a Rademacher distribution, or normally distributed values. The high dimensionality ensures that these randomly generated vectors satisfy the properties described earlier.

\subsubsection{Bundling}
Given a collection of HVs, \emph{bundling} (also called aggregation or superposition) yields a vector similar to all elements in the collection. For example, bundling three vectors is denoted as
\[
\mathbf{u} = [\mathbf{v}_1 + \mathbf{v}_2+\mathbf{v}_3]\,,
\]
where $[\ldots]$ denotes a potential normalization operation. In the case of binary vectors, for example, normalization corresponds to thresholding such that $\mathbf{u}$ is again a binary vector, and aggregation boils down to an element-wise majority rule.
Here, we have that $\mathbf{u}\sim \mathbf{v}_1$, $\mathbf{u}\sim \mathbf{v}_2$ and $\mathbf{u}\sim \mathbf{v}_3$ where `$\sim$' informally denotes that two vectors are similar.
In the case of binary (0/1) or bipolar (-1/1) vectors, `similar' means that they share more elements than expected by chance. 

As an example, consider the task of finding a vector that represents the set of all hydrophobic amino acids. For binary HVs, one could solve the closest string problem, an NP-hard problem that finds the bitstring with the smallest Hamming distance to all the given hypervectors. In practice, however, one often uses a much simpler method: the HVs of the hydrophobic amino acids are bundled using element-wise majority. When bundling an even number of elements, one has to adopt a convention to resolve ties by setting a default value or randomly picking one. Bipolar vectors are particularly easy to bundle, as one can add the vectors and take the sign of the elements; in the case of ties, one can use 0 as a neutral element and upgrade to ternary hypervectors. Taking the average vector for real-valued HVs seems appropriate, though this will reduce the aggregate's norm, violating the homogeneity property. This can easily be understood from the variance rule for independent random variables:
\[
\text{Var}\left[\frac{X_1 + X_2}{2}\right] = \frac{\text{Var}[X_1]+\text{Var}[X_2]}{4}\,.
\]
To bundle $n$ vectors, it is better to either compute $n^{-1/2}\sum_i \mathbf{v}_i$ or to renormalize the sum to match the norm of an atomic HV.

\subsubsection{Binding}

Though powerful, bundling alone cannot represent complex, hierarchical structures. For example, suppose one has the dimer \texttt{AC} (alanine and cysteine) and the dimer \texttt{CE} (cysteine and glutamic acid). In that case, one cannot directly create a bundling from which the identity of these dimers can be recovered. A superposition of both dimers would represent a bag of amino acids, unable to specify which nucleotides are connected to each other in a dimer.  This problem is called the \emph{superposition catastrophe}~\cite{rachkovskijBindingNormalizationBinary2001}. Binding, denoted by $\circ$, solves this problem by generating a new vector from two old ones:
\[
\mathbf{u} = \mathbf{v}_1 \circ \mathbf{v}_2 \,,
\]
such that 
$\mathbf{u} \nsim \mathbf{v}_1$ and $\mathbf{u} \nsim \mathbf{v}_2$, where `$\nsim$' indicates that the vectors are not similar. For bitvectors, element-wise XOR-ing serves well. For bipolar or real-valued HVs, one typically uses element-wise multiplication, though alternative binding operations such as circular convolution~\cite{plate1995holographic} are also used. Importantly, binding is often reversible and does not destroy the information, i.e., there is an \emph{unbinding} operator $\oslash$ that reverses the binding and releases the bound information:
\[
\mathbf{v}_1 \oslash \mathbf{u} = \mathbf{v}_1 \oslash \left( \mathbf{v}_1 \circ \mathbf{v}_2 \right) =  \mathbf{v}_2 \,.
\]
For binary and bipolar HVs, binding and unbinding are the same operations, e.g., XOR is self-inverse. 
Combining bundling and binding allows one to store a data record, i.e., a set of key-value pairs $\mathbf{u}_1 \circ \mathbf{v}_1, \ldots, \mathbf{u}_n \circ \mathbf{v}_n$, which one can query as follows:
\[
\mathbf{u}_i\oslash [\mathbf{u}_1 \circ \mathbf{v}_1+ \ldots+ \mathbf{u}_n \circ \mathbf{v}_n] = [\mathbf{v}_i + \text{noise}]\approx \mathbf{v}_i\,.
\]
The above operation is one of the central ideas behind inference with HVs. For example, by storing a collection of sequences with a bound functional annotation (e.g., enzymes with their associated EC numbers), one can query with new sequences to obtain the likely function annotation. Additionally, this data record encoding is a generic template for encoding different types of data, allowing to store feature identifiers as keys and their associated values.

\subsubsection{Permutation and shifting}
A special case of binding is binding by \emph{permutation}, creating a variant $\rho(\mathbf{v})$ of a single HV $\mathbf{v}$ such that
\[
\rho(\mathbf{v}) \nsim \mathbf{v}\,.
\]
Permutation generates a concept variant, such as the phosphorylation of a protein or the methylation of a nucleotide. The permutation is often implemented as a circular vector shifting with one or more positions, denoted as $\rho^i(\mathbf{v})$. One can easily invert this operation by shifting the corresponding number of positions in the opposite order, i.e., $\rho^{-i}(\mathbf{v})$. Permutation is often used to generate bindings of sequences that retain order information. For example, one can embed the amino acid sequence \texttt{GNP}
\[
\rho^1(\mathbf{v}_G) \circ \rho^2(\mathbf{v}_N)\circ\rho^3(\mathbf{v}_P)
\]
from the respective amino acid HVs.

\subsubsection{Similarity}

The above operations suffice to create arbitrarily complex structures in the hyperdimensional space. One can extract information from this space by comparing vectors based on \emph{(dis)similarity}.
A meaningful similarity measurement is vital for performing \emph{inference}. One often tries to find the entity in the data space that matches the HV result most closely, either by search or optimization. Typically, the large dimensionality ensures that the similarity between two arbitrary HVs is tightly bound, leading to an extremely high sensitivity to detect related HDs.

For bitvectors, one often uses similarities based on the Hamming similarity. The normalized Hamming similarity is given by
\[
    \tau_\text{ham}(\mathbf{u}, \mathbf{v}) = \frac{1}{N}\sum_{i=1}^N \delta_{{u}_i, {v}_i}\,,
\]
with $\delta_{x,y}$ the Kronecker delta function, yielding 1 if $x=y$ and 0 elsewise. This relative Hamming similarity yields values between 0 and 1, with two randomly generated vectors having a value of~0.5. 

In bioinformatics, the Jaccard index (often called the Tanimoto similarity to compare chemometric fingerprints~\cite{Bajusz2015Tanimoto}) is a popular alternative. It is the ratio between the number of elements that equal 1 in both vectors to the number of elements that equal 1 in at least one of the vectors:
\begin{equation}
    \tau_\text{jac}(\mathbf{u}, \mathbf{v}) = 
    \frac{\mathbf{u}\cdot\mathbf{v}}{\mathbf{u}\cdot\mathbf{u} + \mathbf{v}\cdot\mathbf{v}- \mathbf{u}\cdot\mathbf{v}}\,.
\end{equation}
The Jaccard index also gives values in $[0,1]$ with 1/3 the expected value for comparing two random vectors ($0.5^2/(1-0.5^2$)). Since the Jaccard index is appropriate for comparing sets, every position of the HVs is interpreted as a holographic property that the entity does or does not possess, similar to how molecular fingerprints yield information on whether a subgroup is present or absent in a molecule.

For bipolar or real-valued HVs, the cosine similarity is a more natural choice:
\[
    \tau_{\cos}(\mathbf{u}, \mathbf{v}) = 
    \frac{\mathbf{u}\cdot\mathbf{v}}{|\mathbf{u}|\,|\mathbf{v}|}\,.
\]
Here, the output ranges from -1 to 1, and the similarity of two randomly generated vectors is expected to be close to 0. 

\subsection{Encoding of data types}
\label{sec:encodingofdatatypes}
Armed with the four basic operations of HDC, one can map all kinds of objects, such as sequences, graphs, or vectors, into the hyperdimensional space. 
Several strategies exist for all the different data types. As often in data science, some feature engineering might be required to obtain the best representations for a given application.
As a general guideline, similar objects should result in vectors with an increased similarity. We refer to~\cite{kleykoSurveyHyperdimensionalComputing2022} for a more comprehensive survey.

\subsection{The atomic building blocks}
The first step for a given data type is typically identifying the atomic building blocks (e.g., amino acids for protein sequences or proteins for protein-protein networks) and representing them using random generation. Next, these can be combined into structured hierarchical object representations using bundling, binding, and permutation. 

\paragraph*{Symbols} 
Atomic building blocks, such as symbols representing a unique concept, can be generated directly. These symbols might, for example, represent the characters of biological sequences, metabolites, or elements from some ontology. Because of the hyperdimensionality, randomly generated vectors are all dissimilar, meaning that these concepts can be seen as independent. If one wants to encode that one concept is semantically closer to another concept, one can randomly copy a small fraction of one HV to the other, making them more similar~\cite{rachkovskijSparseBinaryDistributed2005}. A more general way of embedding semantic information in HVs is embedding them into a graph where semantically similar concepts are connected and optimizing the HVs to minimize an energy function over this 
graph~\cite{Sutor2018HDCsemantics,Mitrokhin2019HDVrobotpercep}.

\paragraph*{Scalars}
Like nodes in a graph, scalars are another data type where some values are semantically closer to one another than unrelated symbols. For scalars, it is vital to incorporate the notion of closeness. A scalar, representing, for example, expression, is usually represented by considering a fixed range of values divided into discrete bins. The hypervector representing one bin is typically constructed by randomly changing a fraction of components of the hypervector of the previous bin. One can achieve different similarity patterns with varying properties and resolutions by defining the bin width and the number of randomly changed components. The bundling of neighboring bin representations can be interpreted as an approximation of values right in between -- alternatively, more continuous approaches without discrete bins exist~\cite{hernandez2021reghd, dewulf2023hyperdimensionalB}. We refer to~\cite{kleykoSurveyHyperdimensionalComputing2022} for a more detailed overview.

\subsection{Composite objects}

\paragraph*{Numerical objects}
Numerical composite objects, such as real-valued vectors (e.g., gene expression) or functions (e.g., dose-response curves), can be constructed using the aforementioned atomic scalar representations and operations. For example, small vectors can be encoded by binding their scalar elements, likely by shifting to encode the position. Alternatively, to encode a larger vector $\mathbf{x}$, one can use a random projection
\begin{equation}
\mathbf{v} = S \mathbf{x}
\end{equation}
where $S$ is a random, potentially sparse projection matrix containing normally distributed values or elements from $\{-1,1\}$. Some schemes, such as~\cite{imaniBRICLocalitybasedEncoding2019}'s BRIC, suggest a specific structure in the projection matrix to promote hardware optimizations. The resulting HV $\mathbf{v}$ might need to be thresholded, sparsified or normalized. Such random projections are well established with an extensive body of theoretical justification, e.g., the Johnson–Lindenstrauss lemma~\cite{johnsonExtensionsLipschitzMappings1984,Mahoney2011,drineasRandNLARandomizedNumerical2016}, for why they retain the properties of $\mathbf{v}$. Additionally, using well-chosen numerical value encodings, more complex numerical objects such as functions and distributions can be arbitrarily closely approximated using integral transformations~\cite{fradyComputingFunctionsUsing2021,dewulf2023hyperdimensionalA,dewulf2023hyperdimensionalB}.

\paragraph*{Sets and sequences}
Sets can be represented as an aggregation in terms of its symbols. This aggregation acts similarly to a Bloom filter~\cite{kleykoAutoscalingBloomFilter2020,Thomas2021theoryofHDC}, a stochastic data structure used for checking whether an element is part of a set using via multiple hash functions.

Sequences, such as DNA, RNA or peptides, differ from sets in that the order of the symbols matters. Merely bundling the symbols would typically not suffice. To account for the order, one can encode the position using shifting, e.g., no shift for the first symbol in the sequence, a shift of one for the second symbol and so on. One can form the HV of the sequence either by bundling, e.g.,
\begin{equation}
\mathbf{u} = [\rho^0(\mathbf{v}_1) +\rho^1(\mathbf{v}_2)+\rho^2(\mathbf{v}_3)+\rho^3(\mathbf{v}_4)]
\end{equation}
or using binding:
\begin{equation}
\mathbf{u}' = \rho^0(\mathbf{v}_1) \circ \rho^1(\mathbf{v}_2) \circ\rho^2(\mathbf{v}_3)\circ \rho^3(\mathbf{v}_4)\,.
\end{equation}
When using bundling, one can measure the similarity between two sequences based on their representation. An HV obtained by binding the sequence is dissimilar to the representation in which a single symbol differs. When encoding longer sequences, such as proteins or whole genomes, one typically uses the $n$-gram approach (often called $k$-mer in bioinformatics),
using both binding and bundling. Here, one typically represents all subsequences of length $n$ using binding, after which the $n$-gram representations are bundled into one sequence representation.
It might be beneficial to combine several different representations at different levels. For example, to encode a bacterial genome, one might combine multiple $n$-gram representations with a representation based on the presence of the different genes, themselves encoded based on their DNA and protein-coding sequences.

\paragraph*{Graphs}
Graphs, such as metabolic networks, protein-protein networks, or molecules, are also structured datatypes consisting of vertices and edges. Vertices can be atomic or composite. Representations of an edge can be constructed by combining the representations of the corresponding nodes as done in GraphHD~\cite{poduvalGrapHDGraphBasedHyperdimensional2022} and GrapHD~\cite{nunesGraphHDEfficientGraph2022}. One can directly bind the two node vectors if the graph is undirected. When the graph is directed, e.g., in gene-regulatory networks, one can shift one of the node vectors to distinguish between an ingoing and an outgoing edge. When all edges are encoded, they can be bundled to create an HV representing the whole graph. These HV representations allow for solving graph problems such as graph matching, shortest path discovery, graph classification and object detection.

\paragraph*{Images}
Images are the last data type we consider. An image is usually a two-dimensional matrix in which the elements represent the pixel values, either as brightness, color, or something more specialized such as different channels of microscopy images. Again, one can represent the whole image by bundling the pixels with the appropriate spatial context. A simple way to encode this context is by defining two permutation types, representing the pixels' coordinates. This approach has the drawback of not accounting for the closeness between pixels. Role-filler binding-based representations~\cite{rachkovskij2022representation} mediate this problem. Here, close positions are made more similar analogously as for scalars. One effective alternative to creating image representations directly from the pixel values is using a hidden layer of a (convolutional) deep neural network (see, e.g., \cite{yilmazAnalogyMakingLogical2015}). This strategy can also be used for other data types with associated pretrained deep neural architectures, such as (graph) convolutional neural networks or transformers. 

\subsection{Learning with hypervectors}
Here, we give an example of the practical learning flow for machine learning with HDC, depicted in Figure~\ref{fig:operations}(d).
The majority of the work within HDC focuses on classification~\cite{geClassificationUsingHyperdimensional2020,kleykoSurveyHyperdimensionalComputing2022}, but some variants for regression exist~\cite{hernandez2021reghd, frady2022computing}.
Typically, one first maps the data to hypervectors using the methods described in Section~\ref{sec:encodingofdatatypes}. Then, these encoded data points are processed for learning and reasoning using the operations in the hyperdimensional space. Finally, similarity measurements allow for mapping the processed hypervectors back to interpretable predictions.

More concretely, classification, e.g., embedding variants of a particular protein with a function, is typically performed using \emph{prototype methods}~\cite{geClassificationUsingHyperdimensional2020}. Each class has a prototype HV designed such that the classification of a new data point can be performed based on similarity measurement. The predicted class is the one for which the prototype HV is most similar to the HV representation of the data point to predict.

Different heuristic algorithms exist to compute class prototype hypervectors.
The basis is bundling all the HV representations of the members of a class. Although simple bundling is computationally efficient, easy to implement and often works reasonably well, it frequently falls short in predictive performance compared to other contemporary machine learning methods. The predictive performance can be greatly improved by various \emph{retraining} algorithms.
Typically, one cycles through the training set several times, during which wrongly classified examples are added to the correct class prototype and subtracted from the wrongly associated prototype~\cite{geClassificationUsingHyperdimensional2020, hernandez-canoOnlineHDRobustEfficient2021}. 
For example, assume two classes $A$ and $B$ with initial hypervectors $\mathbf{C}_A$ and $\mathbf{C}_B$ obtained by bundling. If a data point, represented by hypervector $\mathbf{v}$, is misclassified as $A$, then one updates $\mathbf{C}_A \leftarrow \mathbf{C}_A - \alpha \mathbf{v}$ and $\mathbf{C}_B \leftarrow \mathbf{C}_B + \alpha \mathbf{v}$ with $\alpha$ the learning rate. Different variants with, e.g., data-dependent or iteration-dependent learning rates, exist to increase performance or speed of convergence~\cite{imani2019adapthd}.

\section{Strengths of HDC for bioinformatics}\label{sec:strengths}

 \begin{figure}
    \centering
    \includegraphics[width=\textwidth]{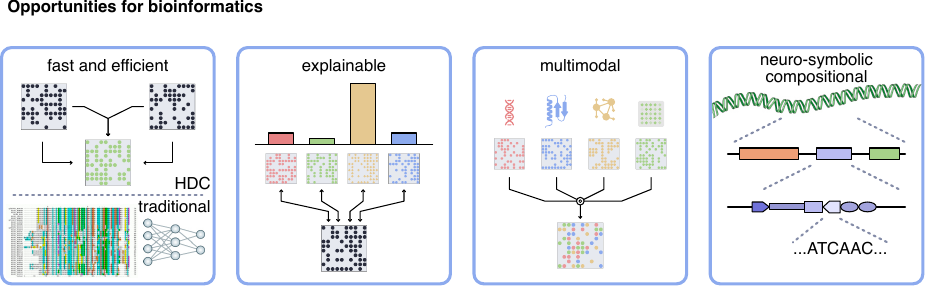}
    \caption{Opportunities for bioinformatics. i) HDC is computationally efficient because it can usually be done using simple bit or arithmetic operations, ii) it is explainable because of its reversibility, iii) it can easily combine different types of data sources, and iv) it can represent complex, structured and hierarchical information.
   }
    \label{fig:opportunities}
\end{figure}

Though HDC is gaining some prominence, it remains relatively underexplored for bioinformatics applications compared to other machine learning approaches. The HDC paradigm can be instrumental in bioinformatics because the field increasingly deals with large amounts of sequence data linked with knowledge. More specifically, in this work, we identify four opportunities that HDC can bring to the field of bioinformatics (Figure \ref{fig:opportunities}):
\begin{enumerate}
\item \textbf{fast and efficient}: HDC has the potential to be much faster than classical alignment algorithms or DL approaches;
\item \textbf{explainable}: the operations in HDC are tractable and often reversible, making them close to white box operations;
\item \textbf{multimodal}: all data are mapped to the same $N$-dimensional vectors, combining different sources of data (e.g., transcription and metabolomics or sequence and structure) is trivial;
\item \textbf{symbolic and hierarchical}: HDC is equipped with an algebra to reason about structured data, such as representing a gene construct.
\end{enumerate}
A similar set of strengths was explored by~\cite{rahimi2018efficient} in the context of biosignal processing.

The field of bioinformatics is generating ever-growing amounts of data~\cite{pal2020big}.
This is primarily due to plummeting sequencing costs, making reading expression possible from the individual cell level to the level of entire microbial communities. In addition, other rich and diverse data sources, such as metabolomics~\cite{plekhova_rapid_2021}, imaging~\cite{haase2022image}, and flow cytometry~\cite{lei2018high}, are also becoming available in a high-throughput fashion.
These data types require specific data processing algorithms to be analyzed and compared.
For example, sequence analysis is driven by advancements in sequence alignment algorithms.
Machine learning methods, and specifically DL architectures, represent flexible, trainable operations with unprecedented power -- and often exorbitant computational demands!
Hyperdimensional computing is seen as a time- and energy-efficient form of machine learning because processing is extremely fast despite the large size of the vectors.
The reason is that encoding, training and inference usually require only simple element-wise operations. 
The operations can often be done using bitvectors, allowing for efficient low-level encoding.
Due to the simplicity of the operations, HDC systems can be implemented on specialized hardware, such as GPUs~\cite{simonHDTorchAcceleratingHyperdimensional2022,al-hashimiTuringNeumannComputational2023} and FPGAs~\cite{imaniSemiHDSemiSupervisedLearning2019,salamatAcceleratingHyperdimensionalComputing2020}.
For example, Demeter~\cite{shahroodiDemeterFastEnergyEfficient2022}, an HDC-based metagenomics profiler, used extensive hardware optimizations to attain more than a hundred-fold speed improvement and thirty-fold memory improvement compared to Kraken2\cite{wood2019improved} and MetaCache~\cite{muller2017metacache}, while maintaining comparable accuracy.

There is a large gap between the vast amount of data and the generation of biological knowledge.
Ideally, a model must be explainable to create robust predictions so the user can verify its assumptions, i.e., why something is predicted and not something else~\cite{hanChallengesExplainableAI2022}.
Many approaches exist toward explainable machine learning~\cite{Molnar2019}, either as models that are naturally interpretable or using posthoc analysis such as Shapley value analysis~\cite{strumbeljExplainingPredictionModels2014}.
Symbolic regression methods can directly distill parsimonious, human-readable rules from data, often with great accuracy~\cite{Wilstrup2021symbolicregressionsmalldata,cranmer2023interpretable}.
Given that HDC works with large, randomly constructed high-dimensional vectors, it is surprising that it is quite explainable.
This is because HDC offers reversible operations, meaning one can decompose complex representations to learn how they work.
One can use similarity matching to compare the HV with different components to see what is essential. For example, to learn which groups or combinations of groups are responsible for the biological activity of a molecule. 

Bioinformaticians not only have to deal with more data, but these data are also becoming more diverse.
Data fusion combines data from different modalities that provide separate and complementary views on common phenomena to solve an inference problem.
Considering the different data sources to discover molecular mechanisms, sample clustering, or attaining the best predictions is far from trivial~\cite{bersanelliMethodsIntegrationMultiomics2016}.
In precision medicine, for example, one can describe a patient's health status using various omics, metabolites and biomarkers, the microbiome, wearable reading, and the environment~\cite{stahlschmidtMultimodalDeepLearning2022}.
Deep learning has shown considerable success in data fusion, as the hierarchical representation makes such models very suitable for multimodal learning~\cite{stahlschmidtMultimodalDeepLearning2022}.
In HDC, different data sources are mapped to the same vector types, bringing them to equal footing.
Simple binding or more complex strategies can combine the different vectors into a single HV representing the different modalities of the object.
For example, a fusion of different types of wearable sensor reading -- electroencephalography recordings, accelerometers, galvanic skin response -- can accurately detect human activity and emotions~\cite{changHyperdimensionalComputingbasedMultimodality2019,zhaoAttentiveMultimodalLearning2023}. 

A final aspect where HDC shines is representing complex, structured hierarchical information. 
Biological data is inherently hierarchical and nested: protein networks consist of proteins, which include domains and amino acids.
The most potent representations also incorporate the aspects of the lower-level constituents.
However, combining complex information is a challenging problem, referred to as the binding problem~\cite{greffBindingProblemArtificial2020}.
For example, combining the concepts of a `red apple' and a `green pear' might lose the specific color-object associations. In bioinformatics, an example would be adding semantic information to individual genes in a set.
The operations of HDC are supremely suited to handle this issue, allowing one to freely combine specific concepts due to the distributivity properties of the operations. For example, in image processing, one can use bundling to combine several holistic image descriptors with local image descriptors of specific regions in the image for more accurate place recognition for mobile robotics~\cite{neubertHyperdimensionalComputingFramework2021}.
This allows HDC to thrive for problems with reasoning and structure, such as recently outstripping DL on Raven's progressive matrix problem~\cite{herscheNeurovectorsymbolicArchitectureSolving2023}.

Many tricky machine learning problems can become trivial using HDC.
For example, suppose one has computed several energetically favorable RNA secondary structures, but one does not exactly know which one(s) is or are biologically active -- an instance of multi-instance learning~\cite{fatimaComprehensiveReviewMultiple2023}.
Such problems are ubiquitous in bioinformatics. They are easily handled by aggregating to obtain an HV similar to all candidates.

\section{Opportunities for bioinformatics}\label{sec:opps}

Here, we identify several domains in bioinformatics in which HDC has proved valuable. In addition, we also speculate on which other domains in bioinformatics remain to be explored and which domains might not be a perfect fit for HDC.

Firstly, HDC has proved its worth in processing omics data. Within the omics domain, problems usually involve matching high-throughput generated data to a reference database. This application is especially useful for HDC, given its speed and low memory footprint. Notably, because HDC works with fixed-shape representations, (sub)sequence matching becomes independent of the length of the reference sequence. Several studies have reported magnitudes of improvements in both speed and energy use compared to the state-of-the-art. HDNA~\cite{imaniHDNAEnergyefficientDNA2018}, GenieHD~\cite{kimGenieHDEfficientDNA2020}, BioHD~\cite{zouBioHDEfficientGenome2022} and HDGIM~\cite{barkamHDgim2023} are HDC-based frameworks to match DNA sequences to reference databases efficiently. Often, these make use of highly parallelized implementations and specific hardware optimizations. For example, BioHD uses processing in memory (PIM) for massive parallelism to obtain 100$\times$ speedups and energy efficiency, even compared to other algorithms running on PIM accelerators. As a tool for protein back-translation, they resolve ambiguities in similar-encoding nucleic acid sequences by superposing them. Another example is Demeter~\cite{shahroodiDemeterFastEnergyEfficient2022}, a metagenomics profiler made for real-time monitoring of food. The authors use specific memristor optimizations to obtain large memory reductions and speed improvements compared to state-of-the-art methods while seeing only negligible drops in accuracy. In epigenetics, HDC was successfully used to classify tumor and non-tumor sequences based on their methylation profile~\cite{Cumbo2020methylationHDC}.
Alternative sequence encodings were proposed, improving performance on tasks such as protein secondary structure prediction~\cite{rachkovskij2021shift}. These encodings provided equivariance concerning the shift of sequences and preserved the similarity of sequences with identical elements.

A second domain with large amounts of data relates to biosignals and spectra. Here, HDC can also provide fast ways to analyze large-scale data at competitive performance. For example, HyperSpec~\cite{xuHyperSpecUltrafastMass2023} is a HDC-based approach for clustering mass spectrometry data that achieves speedups of up to 15-fold compared to alternative clustering tools. In addition, HyperSpec combines both the spatial locality of the spectra peaks and the intensity of those peaks, making it an excellent example of HDC's advantage in coping with complex data. HDC has been used for classifying the sensitivity of glioma to chemotherapy using proteomics SELDI-TOF spectra~\cite{rachkovski2010intelligent}. Similarly, the recently proposed HyperOMS~\cite{kangMassivelyParallelOpen2023} is an HDC-based algorithm for open modification spectral searching in mass spectrometry proteomics for identifying post-translational modifications. 

Related to biosignal processing, a prominent example using HDC is the work by~\cite{rahimiRobustEnergyefficientClassifier2016}, who developed a set of HD architectures for encoding ExG signals across multiple modalities and demonstrated how these architectures result in explainable vectors. In later work, \cite{rahimi2018efficient} extensively explored the potential of HDC for various ExG biosignals (i.e., electromyography, electroencephalography, and electrocardiography) and found equal to superior performances compared to the state-of-the-art, while HDC (i) demanded much less data and could work in the zero-shot setting, (ii) dealt well with noisy and unprocessed inputs and (iii) proved to be transparent and repeatable.
In general, a lot of work within HDC has focused on the processing of bio(medical)signals, often on IEEG, EEG, or EMG signals, for example, in seizure detection, septic shock modelling, and hand gesture recognition~\cite{menon2022highly,kleyko2019hyperdimensional,schindlerPrimerHyperdimensionalComputing2021,moin2021wearable,moin2018emg,burrello2020ensemble, benatti2014analysis, ge2022applicability,watkinsonClassModelingSepticShock2021,lazarou2018eeg,rahimi2018efficient,zou2022eventhd,rahimi2016hyperdimensional,pale2022hyperdimensional,rahimi2020hyperdimensional,burrello2019hyperdimensional,basaklar2021hypervector,zhou2021incremental,burrello2019laelaps,zhang2015low,ni2022neurally,burrello2018one,benatti2019online,pale2021systematic,kleyko2018vector}.

Thirdly, HDC methods are suitable for learning with molecules and graphs. Graphs and networks are invaluable tools in systems biology, for example, in metabolic networks, protein-protein networks or gene regulatory networks. 
GraphHD~\cite{nunesGraphHDEfficientGraph2022} and GrapHD~\cite{poduvalGrapHDGraphBasedHyperdimensional2022} are general HDC-based approaches for encoding and classifying graphs, achieving comparable performance as state-of-the-art methods on real-world classification problems. HyperRec~\cite{guo2021hyperrec} is a recommender system based on HDC. The method encodes items and users into hypervectors to predict rankings for new items based on user preferences, which can be seen as predicting links in a graph. 
Graphs containing drug-drug, protein-protein, and drug-target interactions were represented as hypervectors to predict new adverse drug-drug effects~\cite{burkhardt2019predicting}.
Also, the hierarchical structure of atoms in a molecule represents a graph and can used to predict molecular properties based on HDC~\cite{slipchenko2005distributed}.
MoleHD~\cite{maMoleHDEfficientDrug2022} is a more recently proposed HDC tool for molecular property prediction, such as permeability through the blood-brain barrier or drug side-effects. This method performs favorably compared to several state-of-the-art methods, including graph convolutional neural networks, while requiring mere minutes to train on a CPU.

HDC is also being applied in the domain of online health care. It is an excellent fit due to its efficiency and multimodal-friendly characteristics. For example, \cite{changHyperdimensionalComputingbasedMultimodality2019} developed HDC-MER, a HDC framework for emotion recognition based on multiple modalities that are encoded into hypervectors across time. \cite{menon2022highly} also developed an HDC method for emotion recognition but specifically leveraged a cellular automaton for hypervector generation to maximize energy efficiency in settings with many modalities. A third example is the work by \cite{watkinsonDetectingCOVID19Related2021}, in which a proposed HDC method can analyze computed tomography scans for early COVID-19 detection. Furthermore, HDC has been applied for seizure~\cite{schindlerPrimerHyperdimensionalComputing2021} and septic shock detection~\cite{watkinsonClassModelingSepticShock2021}, in overlap with the works cited earlier on the processing of biomedical signals. 

HDC was used for natural language processing and analogical retrieval of information using the predication-based semantic indexing~\cite{cohen2012many,cohen2009empirical}. For example, from the composition in the sentences such as ``drug A \emph{treats} disease B'', one may infer the predicate pathway "drug A \emph{interacts with} gene C \emph{associated with disease B"}. In this way, HDC can be used to mediate the identification of therapeutically valuable connections for literature-based discovery~\cite{cohen2012discovering}. Similarly, HDC was discussed in the context of pharmacovigilance, drug repurposing, and discovery-by-analogy~\cite{cohen2017embedding,cohen2014predicting}. Note that these (analogical) interferences of interactions based on language show a strong connection to the more explicit graph representations discussed earlier, emphasizing the multi-modality of HDC. The scale at which modern DL techniques can process large, diverse datasets has given rise to \emph{foundation models}, general models capable of being adapted to a wide range of downstream tasks~\cite{bommasaniOpportunitiesRisksFoundation2022}. Large models, such as BioBERT~\cite{leeBioBERTPretrainedBiomedical2020}, allow for the processing of large amounts of biomedical data, for example, for drug discovery or personalized medicine~\cite{guDomainSpecificLanguageModel2021}. HDC can complement such approaches as its strengths complement the weaknesses of DL, i.e., HDC systems being lightweight to train and deploy and their transparent operations. Here, HDC is suited to create relatively small, highly specialized knowledge systems.
Finally, HDC is also used in medical imaging~\cite{kleyko2017modality}. For example, \cite{billmeyerBiologicalGenderClassification2021} used fMRI images for biological gender classification and \cite{watkinson2021detecting} used CT scans for detecting COVID-19-related pneumonia.

Beyond areas where HDC is already being applied, we also see opportunities to apply HDC in yet-to-be-explored areas. Phylogeny is the first domain where HDC would shine. It allows for incorporating various data sources -- genomics, expression, morphology -- in simple vectors that can be compared. The hierarchical nature of HDC would allow one, for example, to encode all the gene variants of a species and combine these in a HV that represents their relative order in the genome. This would allow for studying organisms with complex mosaic genomes, such as phages~\cite{dionPhageDiversityGenomics2020}. 


A final application in which we see a lot of potential for HDC is genetic engineering, biotechnology and breeding. Such endeavors are often very specialized projects, frequently of a proprietary nature, in which a lot of domain knowledge and experimental data are available. For example, enzyme engineering combines wild-type sequence data in their biological context with various mutation experiments and activity and stability assays. This information has to be integrated into a model that correctly incorporates the causal mechanisms so that the most promising new mutations can be highlighted. Synthetic biology is modular in the sense that the basic parts, genes, protein domains or cells, can be combined into new functional entities\cite{purnick2009second}. The composability of HDC can be suitable to represent such designs.

Some application areas seem to be less relevant for HDC. These are areas where highly complex relationships need to be learned with rather limited knowledge and where one can rely on adequate objective functions for optimizing a complex black-box function: cases where DL shines. One example is protein structure prediction, in which the goal does not match the strengths HDC can offer. A second area is generative applications such as protein design. Although HDC methods can be generative, we believe that the goal of protein design needs to be aligned better with the strengths of HDC frameworks. In general, DL's strength is in learning a mapping from one space to another, given that these spaces are densely populated with examples. HDC, however, shines when there is a specific, known structure that one wishes to encode.

\section{Conclusions}

A key idea in bioinformatics is that statistically meaningful similarities indicate a biological signal, a reasoning often based on evolutionary principles. Many alignment-based algorithms for sequences, structures, or graphs exploit this principle by searching large databases for homologs and the like. More recent approaches based on machine learning, specifically deep learning, have been highly successful at learning general maps from complex input to output domains, such as sequence to structure. Their power and generality have transformed nearly every subdomain of bioinformatics.

This work discussed HDC as an additional tool in the bioinformatician's arsenal. HDC shares similarities with the search-based and learning-based paradigms. HDC's strengths nicely complement some of the weaknesses of DL (and likely vice versa). Initially, the most prominent selling point of HDC appears to be its speed and computational efficiency, allowing for training on simple hardware and performing online inferences at scale on specific hardware such as FPGAs~\cite{kleykoVectorSymbolicArchitectures2022}. Future ``unconventional computation'' strategies might use alternative physical, chemical, or biological processes for computation, such as optics~\cite{huangProspectsApplicationsPhotonic2022}, reaction-diffusion processes~\cite{adamatzkyReactionDiffusionComputing2012}, or plants~\cite{pietersLeveragingPlantPhysiological2022}. Hyperdimensional computing would be well suited for such forms of stochastic cybernetic modes of computation~\cite{jaeger2023toward}.

The ability of HDC for structural compatibility using powerful operators might be even more useful than its computational efficiency. These operations allow the bioinformatician to encode prior domain knowledge and the problem structure in the system. Using problem structure is especially important when using the model for guiding interventions, for example, in precision medicine and genetic engineering. The most exciting advancements will likely occur by combining the general, gradient-based mappings of DL with the symbolic reasoning of HDC into neuro-symbolic AI~\cite{smolenskyNeurocompositionalComputingCentral2022}. Recent work highlighted the potential of such hybrids~\cite{mitrokhinSymbolicRepresentationLearning2020,herscheNeurovectorsymbolicArchitectureSolving2023} for combining the compositional power of symbolic reasoning with the flexibility of gradient-based learning.

\section{Funding}
M.V.H. received PhD funding by Research Foundation – Flanders (FWO) – FWO-SBO Bisceps project (grant no. S007019N) and the Ghent University special research fund (BOF.PDO.2024.0003.01). D.B. has received funding from the Research Foundation – Flanders (FWO), grant number 1S69520N. S.T. received funding from the Flemish Agency for Innovation and Entrepreneurship (VLAIO) grant number HBC.2020.2292. P.D. and B.D.B. received funding from the 
Flemish Government under the "Onderzoeksprogramma Artificiële Intelligentie (AI) Vlaanderen" program.

\bibliographystyle{acm}

\end{document}